\documentclass[letterpaper]{article} 
\usepackage{aaai2027}  
\usepackage[hyphens]{url}  
\usepackage{graphicx} 
\usepackage{amssymb}
\usepackage{natbib}  
\usepackage{caption} 
\usepackage{newfloat}
\usepackage{listings}

\usepackage{booktabs}

\title{AgroBench: A Reproducible Multimodal Benchmark for Weakly Supervised Crop Yield Learning from County Statistics and Pixel Observations}

\author{
    Udaiveer Singh\textsuperscript{\rm 1},
    Rajiv Ranjan\textsuperscript{\rm 1},
    Shashank Tamaskar\textsuperscript{\rm 1},
    Dharmendra Saraswat\textsuperscript{\rm 2}
}

\affiliations{
    \textsuperscript{\rm 1}Plaksha University, Mohali, Punjab, India\\
    \textsuperscript{\rm 2}Purdue University, West Lafayette, Indiana, USA\\
    \{udaiveer.singh.ug23, rajiv.ranjan, shashank.tamaskar\}@plaksha.edu.in,\\
    saraswat@purdue.edu
}

\begin{document}

\maketitle

\begin{abstract}

Reliable agricultural yield statistics are typically reported at coarse administrative scales, whereas modern geospatial machine learning methods require spatially explicit, pixel-level supervision. This mismatch has limited the development of large-scale benchmarks for crop yield learning using multimodal Earth observation data. A reproducible benchmark, AgroBench, is presented for transforming publicly available U.S. county-level crop yield statistics into weakly supervised pixel-level crop time series. Each crop-pixel time series is paired with a county-level yield value as a weak supervisory signal rather than a directly measured pixel-level yield label. Our geospatial data generation pipeline integrates USDA crop yield statistics with crop-specific land cover masks, Sentinel-2 multispectral imagery, Sentinel-1 synthetic aperture radar observations, climatic variables, and terrain information to produce temporally aligned multimodal sequences describing individual crop pixels throughout the growing season. The resulting benchmark contains over 13 million observations from 788,654 unique crop pixels spanning 5,107 county-year combinations across eight growing seasons (2017–2024) for five major U.S. crops. To facilitate standardized evaluation, we establish a crop yield prediction benchmark using a Leave-One-Year-Out evaluation protocol and provide baseline results using representative machine learning models. By releasing the complete data generation pipeline, benchmark dataset, and evaluation protocol, AgroBench provides a reproducible foundation for future research in weakly supervised learning, multimodal remote sensing, spatiotemporal modeling, and geospatial foundation models for agriculture.

\end{abstract}


\section{Introduction}
\label{sec:introduction}

Advances in Earth observation have transformed agricultural monitoring through continuous observation of croplands from space. High-resolution satellite imagery, together with recent progress in machine learning and geospatial foundation models, has enabled new approaches for crop yield estimation, crop health assessment, and large-scale agroecosystem analytics. Multimodal observations from Sentinel-2~\cite{sentinel2}, Sentinel-1~\cite{sentinel1}, and complementary environmental datasets provide a rich foundation for learning spatial and temporal patterns of crop development.

Despite these advances, reliable crop yield measurements are rarely available at the spatial resolution required by modern machine learning models. The most comprehensive agricultural statistics, such as those reported by the United States Department of Agriculture (USDA) National Agricultural Statistics Service (NASS)~\cite{usda_nass,usda_quickstats}, are typically published at coarse administrative levels (e.g., counties), whereas geospatial learning models operate on pixels, image patches, or spatially explicit temporal sequences. This mismatch between coarse supervisory signals and fine-resolution Earth observation data remains a major bottleneck for developing large-scale crop yield benchmarks.

Existing public datasets address related agricultural tasks but not this challenge. CropHarvest~\cite{cropharvest}, PASTIS~\cite{pastis}, Sen4AgriNet~\cite{sen4agrinet}, and Sickle~\cite{sickle} primarily support crop classification or segmentation, while SSL4EO-S12~\cite{ssl4eo} focuses on self-supervised representation learning. CropNet~\cite{cropnet} and CYBench~\cite{cybench} provide crop-yield datasets but remain aggregated at administrative or field scales, limiting their suitability for pixel-level learning. A detailed comparison is provided in the section ``Comparison with Existing Datasets.''

AgroBench is a reproducible benchmark for weakly supervised pixel-level crop yield learning that bridges county-level agricultural statistics with multimodal Earth observation data through a scalable geospatial processing pipeline. AgroBench enables large-scale weakly supervised crop yield learning through more than 13 million multimodal observations from 788,654 crop pixels spanning five major U.S. crops across eight growing seasons (2017--2024).

The primary contributions of this work are summarized as follows:

\begin{itemize}

\item A reproducible weak supervision pipeline that bridges the mismatch between county-level crop yield statistics and pixel-level Earth observation data, enabling large-scale benchmark construction entirely from publicly available resources.

\item AgroBench, a large-scale multimodal benchmark comprising more than 13 million observations from 788,654 crop pixels spanning 5,107 county-year combinations, five major U.S. crops, and eight growing seasons, integrating optical, SAR, climate, and terrain observations with weakly supervised crop yield labels.

\item Standardized benchmark protocols and reproducible baseline evaluations under a Leave-One-Year-Out framework, establishing reference performance for future spatiotemporal, multimodal, and geospatial foundation models.

\end{itemize}

\paragraph{Code and Data Availability.}
The AgroBench benchmark generation pipeline, benchmark metadata,
documentation, baseline implementations, and evaluation scripts are publicly
available at \url{https://github.com/udaiveersingh/AgroBench}. An archived
release is available at \url{https://doi.org/10.5281/zenodo.21629503}.
\section{Related Work}
\label{sec:related_work}

\subsection{Earth Observation Foundation Datasets}

Large-scale Earth observation datasets have accelerated self-supervised representation learning for remote sensing. SSL4EO-S12~\cite{ssl4eo} provides multimodal Sentinel-1 and Sentinel-2 imagery for learning transferable representations across diverse Earth observation tasks. However, it does not provide crop yield supervision or standardized benchmark tasks for agricultural prediction.

\subsection{Agricultural Remote Sensing Datasets}

Several public benchmarks support agricultural remote sensing tasks. CropHarvest~\cite{cropharvest} provides field-level labels for crop classification, PASTIS~\cite{pastis} focuses on parcel-level segmentation, Sen4AgriNet~\cite{sen4agrinet} supports crop monitoring, and Sickle~\cite{sickle} provides multi-temporal semantic segmentation. While these datasets have advanced agricultural representation learning, they are not designed for crop yield prediction using weakly supervised pixel-level observations.

\subsection{Crop Yield Prediction Datasets}

Public crop yield datasets remain comparatively limited. CropNet~\cite{cropnet} and CYBench~\cite{cybench} integrate remote sensing with crop yield statistics to support large-scale yield prediction, but their supervisory labels remain associated with administrative or field-level units rather than individual crop pixels. Consequently, they do not support learning spatially explicit multimodal representations under weak supervision.

AgroBench directly addresses this gap by linking county-level crop yield statistics with pixel-level multimodal Earth observation data through a reproducible weak supervision pipeline.

\section{AgroBench Data Generation Pipeline}
\label{sec:pipeline}

The proposed AgroBench dataset is curated through a reproducible multi-stage geospatial pipeline that transforms publicly available agricultural statistics and Earth observation data into weakly supervised pixel-level learning samples. The pipeline integrates county-level crop yield statistics, crop-specific land cover maps, multimodal satellite observations, and environmental variables without requiring new field measurements.

This formulation establishes a standardized benchmark for learning under
label-resolution mismatch, where supervision is available only at coarse
administrative scales while prediction operates on spatially explicit
multimodal observations.

\begin{figure*}[t]
    \centering
    \includegraphics[width=\textwidth]{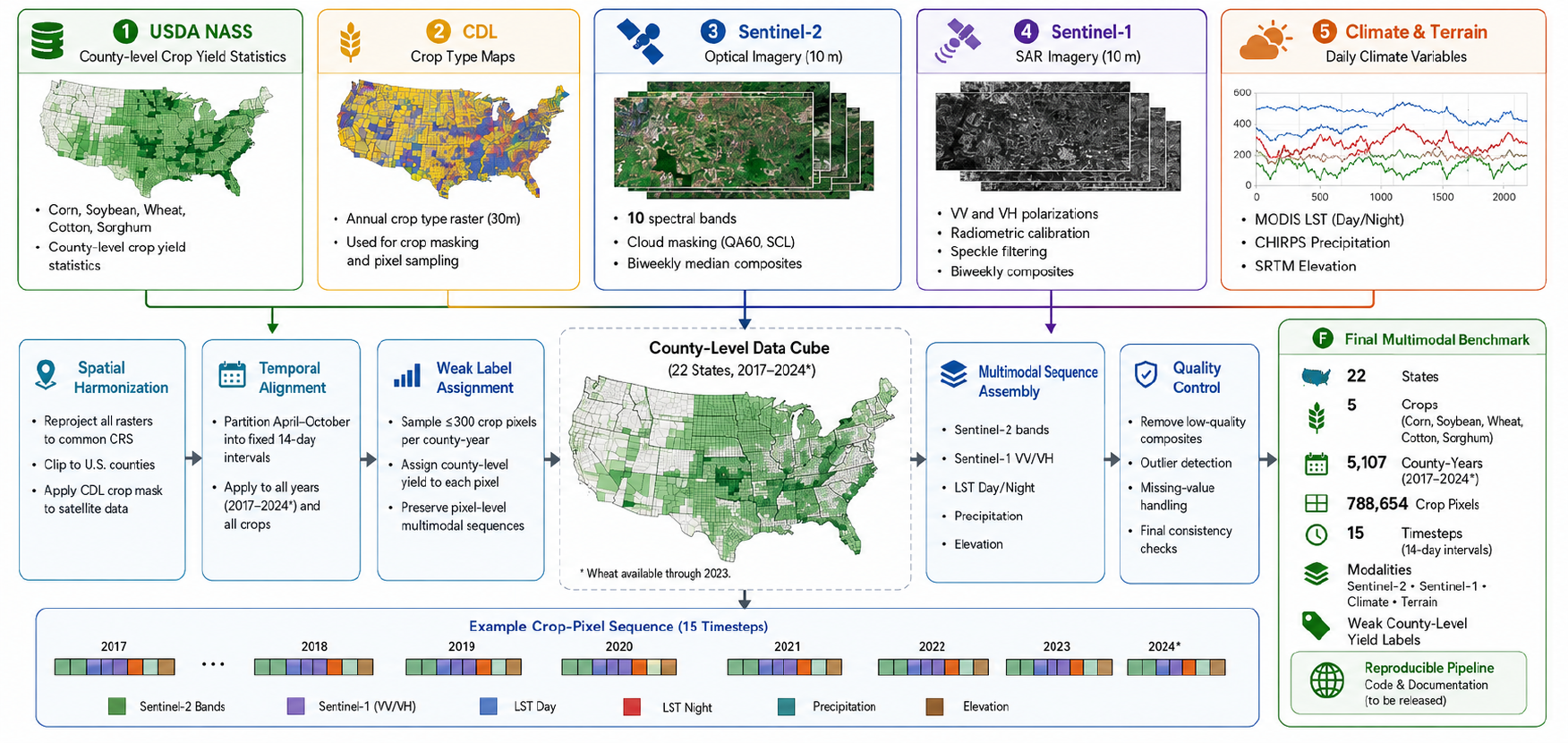}
    \caption{
        Overview of the AgroBench data generation pipeline.
        (1) County-level yield statistics are collected from USDA NASS Quick Stats~\cite{usda_quickstats}.
        (2) Crop pixels are identified using the USDA Cropland Data Layer (CDL)~\cite{usda_cdl}.
        (3) Multimodal Sentinel-1~\cite{sentinel1}, Sentinel-2~\cite{sentinel2}, MODIS LST~\cite{modis_lst}, CHIRPS~\cite{chirps}, and SRTM~\cite{srtm} data are retrieved.
        (4) Observations are temporally harmonized into fixed 14-day intervals.
        (5) Weakly supervised pixel-level samples are assembled.
        (6) Quality assurance produces the final AgroBench benchmark.
    }
    \label{fig:pipeline}
\end{figure*}

Figure~\ref{fig:pipeline} summarizes the six stages of the workflow: (1) county-level yield acquisition, (2) crop pixel identification, (3) multimodal data retrieval, (4) temporal harmonization, (5) benchmark assembly, and (6) quality assurance. The modular design enables straightforward extension to additional crops, regions, and years.

\subsection{Acquisition of County-Level Yield Statistics}
\label{subsec:yield_statistics}

The proposed dataset uses county-level crop yield statistics, obtained from the United States Department of Agriculture (USDA) National Agricultural Statistics Service (NASS)~\cite{usda_nass}, which provides annual county-level crop yield estimates across the United States.

County-level records for corn, soybeans, wheat, cotton, and sorghum were collected from the USDA Quick Stats database~\cite{usda_quickstats}. Data from 2017--2024 were retained for counties with substantial production of each crop.


\subsection{Generation of Weakly Supervised Pixel-Level Labels}
\label{subsec:weak_supervision}

County-level yield statistics provide reliable supervision, whereas modern learning methods operate on individual pixels. AgroBench bridges this resolution mismatch by assigning county-level yield labels to representative crop pixels within each county-year.

Formally, let $y_{c,t}$ denote the reported crop yield for county $c$
during growing season $t$, and let
$\mathcal{X}_{c,t}=\{x_{i,c,t}\}_{i=1}^{N_{c,t}}$
represent the set of sampled crop-pixel sequences within that
county-year, where each $x_{i,c,t}$ is a multimodal temporal
observation. Under the weak-supervision setting, every sampled pixel
inherits the county-level yield label $y_{c,t}$ during training. Pixel-level
predictions $\hat{y}_{i,c,t}$ are aggregated to obtain the county-level
estimate

\[
\hat{y}_{c,t}
=
\frac{1}{N_{c,t}}
\sum_{i=1}^{N_{c,t}}
\hat{y}_{i,c,t},
\]

which is compared against $y_{c,t}$ using the evaluation metrics
described in Section "Benchmark Protocol and Baseline Results".

County boundaries were identified using Federal Information Processing Standards (FIPS) codes, and crop-specific pixels were extracted from the USDA Cropland Data Layer (CDL). Restricting supervision to crop pixels reduces label noise arising from heterogeneous land cover.

To maintain computational scalability while preserving broad spatial representation, up to 300 randomly selected valid crop pixels were retained from each county-year after crop masking. Counties containing fewer than 300 valid crop pixels retained all available pixels, whereas larger counties were capped at 300 samples to avoid disproportionate representation in the final dataset. Each sampled pixel inherits the county-level yield label while retaining its own location and temporal observations, enabling weakly supervised pixel-level learning.

\subsection{Retrieval of Multimodal Earth Observation Data}
\label{subsec:eo_retrieval}

Each sampled crop pixel was enriched with optical, radar, climatic, and terrain observations describing vegetation dynamics and environmental conditions throughout the growing season. Sentinel-2 surface reflectance (10 bands)~\cite{sentinel2} and Sentinel-1 VV/VH backscatter~\cite{sentinel1} were retrieved for each sampled pixel. Missing optical observations caused by cloud cover were represented using placeholder values. Environmental variables included MODIS land surface temperature~\cite{modis_lst}, CHIRPS precipitation~\cite{chirps}, and SRTM elevation~\cite{srtm}, providing complementary climatic and terrain information.



\subsection{Temporal Harmonization Across Sensing Modalities}
\label{subsec:temporal_harmonization}

The Earth observation products incorporated in the proposed dataset differ considerably in their temporal acquisition frequency and revisit characteristics. While Sentinel-2 observations are affected by cloud cover and irregular acquisition schedules, Sentinel-1 provides weather-independent radar observations, MODIS land surface temperature is available as an 8-day product, and CHIRPS precipitation is reported daily. Directly combining these heterogeneous observations would result in temporally misaligned feature sequences that complicate downstream machine learning.

To establish a consistent temporal representation, each growing season was partitioned into consecutive 14-day intervals spanning April through October. Within each interval, optical and radar observations were aggregated, land surface temperature was averaged, precipitation was accumulated, and elevation remained constant. This harmonization produces a fixed-length sequence of 15 temporally aligned observations for every sampled crop pixel, regardless of differences in sensor revisit frequency or temporary data gaps. Each timestep therefore captures aligned optical, radar, climatic, and terrain information corresponding to approximately the same stage of crop development.

\subsection{Construction of the Analysis-Ready Benchmark}
\label{subsec:dataset_assembly}

Following temporal harmonization, all sensing modalities were consolidated into a unified spatiotemporal representation. Each sample contains georeferenced optical, radar, climatic, and terrain observations together with the corresponding county-level crop yield label.

The final benchmark is organized in a tabular format in which each row corresponds to a single pixel observation at one timestep. Every record contains ten Sentinel-2 spectral bands, two Sentinel-1 radar channels, daytime and nighttime land surface temperature, accumulated precipitation, elevation, geographic coordinates, temporal metadata, crop identity, county identifier, and the corresponding weak supervisory signal. This schema preserves the complete temporal evolution of every sampled pixel while maintaining explicit links to its administrative yield statistics.

Quality assurance procedures verified structural consistency, temporal completeness, missing values, and physical plausibility of all retrieved variables. The resulting benchmark characteristics are summarized in the following section.

\begin{table}[t]
\centering
\small
\setlength{\tabcolsep}{4pt}
\caption{Spatial and temporal coverage of the proposed benchmark for each crop.}
\label{tab:crop_coverage}
\begin{tabular}{lcccc}
\toprule
Crop & Years & States & County-years & Unique Pixels \\
\midrule
Corn      & 2017--2024 & 8  & 1,713 & 313,598 \\
Soybeans  & 2017--2024 & 11 & 2,092 & 344,812 \\
Wheat     & 2017--2023$^\dagger$ & 12 & 776 & 126,534 \\
Cotton    & 2017--2024 & 5  & 431 & 67,830 \\
Sorghum   & 2017--2024 & 2  & 95 & 14,246 \\
\bottomrule
\end{tabular}

\vspace{1mm}
\footnotesize{$^\dagger$County-level wheat yield statistics for 2024 were unavailable during dataset construction.}
\end{table}
\section{Dataset Characteristics}
\label{sec:dataset_characteristics}

This section summarizes the composition, coverage, and quality of AgroBench, highlighting the benchmark's spatial and temporal diversity, multimodal structure, and data quality.

\subsection{Dataset Overview}
\label{subsec:dataset_overview}

The benchmark comprises multimodal spatiotemporal observations collected over eight growing seasons (2017--2024) for five major U.S. crops: corn, soybeans, wheat, cotton, and sorghum. Each sample corresponds to a georeferenced crop pixel represented by a temporally aligned sequence of optical, radar, climatic, and terrain observations paired with a weak supervisory signal derived from county-level crop yield statistics.

Overall, AgroBench contains more than 13 million observations from 788,654 unique crop pixels spanning 5,107 county-year combinations. Each pixel is represented by a fixed-length sequence of 15 timesteps covering the primary growing season (April--October), providing a standardized representation for spatiotemporal learning.

\subsection{Geographic and Temporal Coverage}
\label{subsec:coverage}

\begin{figure*}[t]
    \centering
    \includegraphics[width=\textwidth]{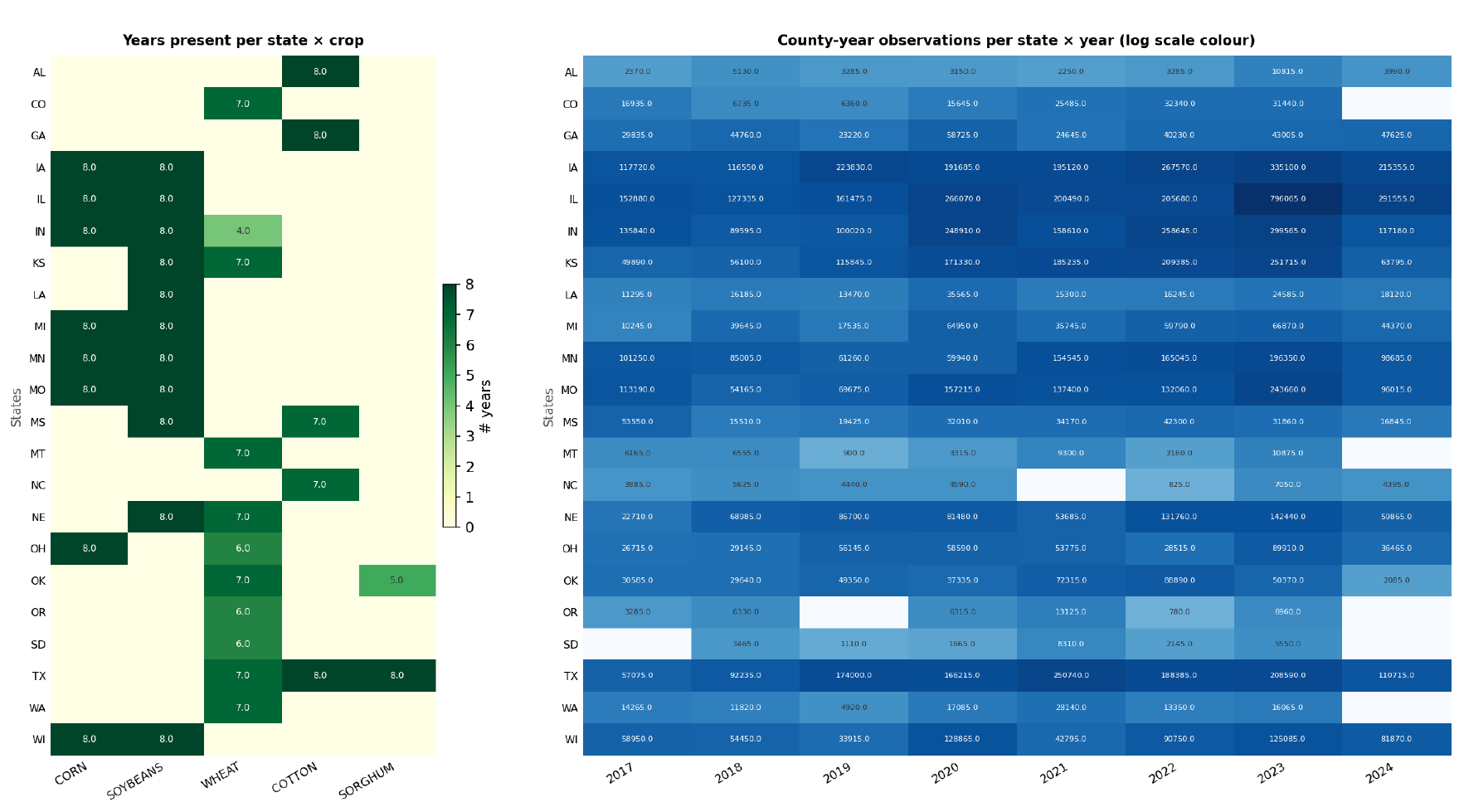}
    \caption{
    Geographic and temporal coverage of the proposed benchmark. Left: availability of yearly observations for each state--crop combination. Right: county-year observations across growing seasons illustrating the spatial and temporal distribution of the dataset.
    }
    \label{fig:coverage_map}
\end{figure*}

AgroBench spans the principal production regions of five major U.S. crops across eight growing seasons (2017--2024), covering 5,107 county-year combinations selected from regions with consistently reported county-level yield statistics. Figure~\ref{fig:coverage_map} summarizes the resulting geographic and temporal coverage.

\subsection{Dataset Composition}
\label{subsec:composition}

\begin{figure*}[t]
    \centering
    \includegraphics[width=\textwidth]{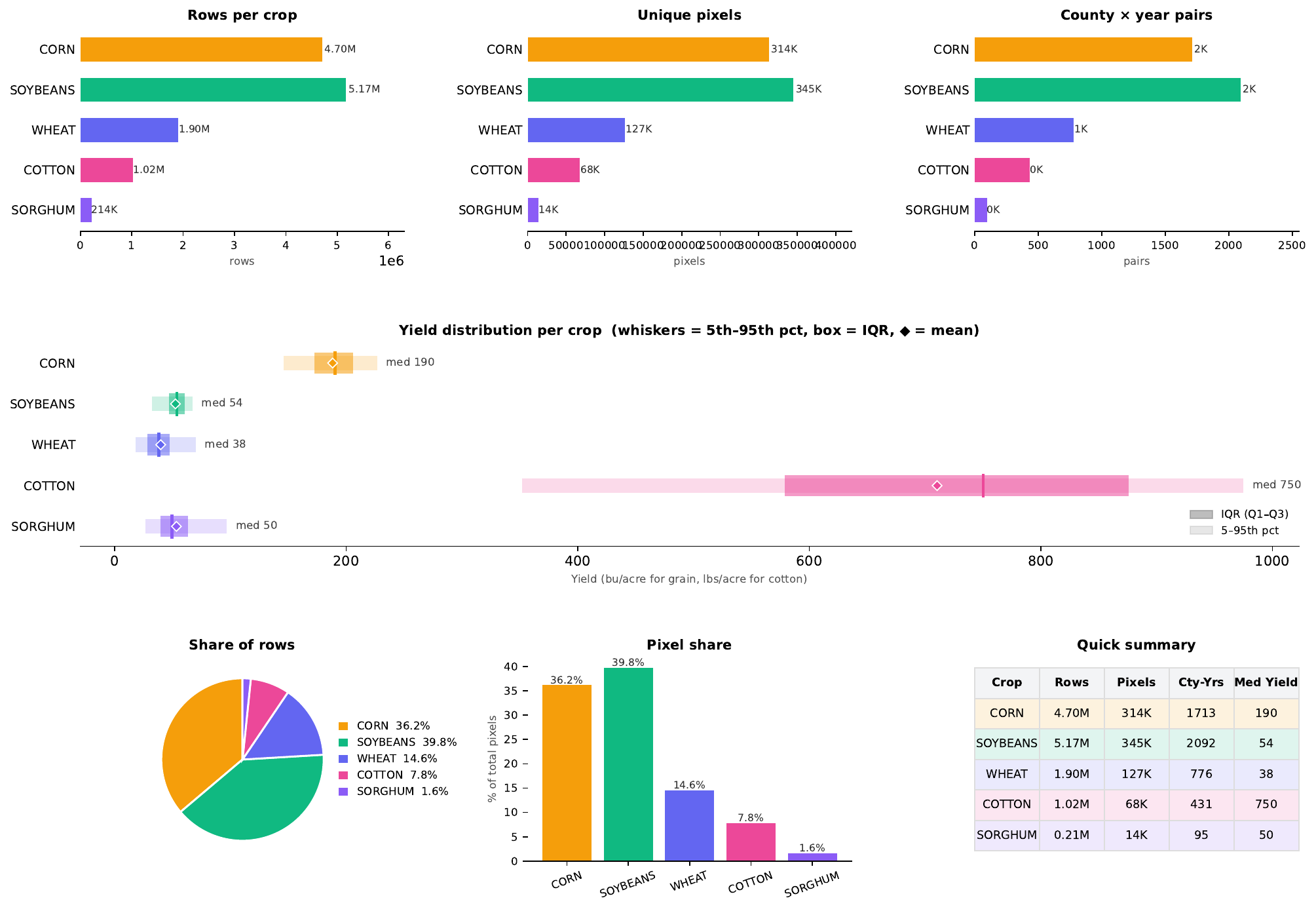}
    \caption{
    Statistical overview of the proposed benchmark, including sample counts, unique pixels, county-year coverage, crop-wise yield distributions, and overall dataset composition.
    }
    \label{fig:distribution}
\end{figure*}

Understanding the composition of AgroBench is important for interpreting its spatial diversity and suitability for downstream machine learning applications. Figure~\ref{fig:distribution} summarizes the distribution of observations, unique pixels, county-year combinations, and crop yield statistics across all five crops.

AgroBench contains more than 13 million temporally resolved observations from 788,654 unique crop pixels spanning 5,107 county-year combinations. Soybeans and corn account for most observations, while wheat, cotton, and sorghum broaden the benchmark across diverse production environments. Together, these crops capture substantial variability in geographic coverage and yield distributions, providing a diverse benchmark for weakly supervised crop yield learning.

\subsection{Data Quality and Validation}
\label{subsec:data_quality}

To ensure the reliability of the proposed benchmark, several quality assurance procedures were performed after dataset generation. These analyses evaluate data completeness, physical consistency of the retrieved variables, and whether the assembled spatiotemporal observations exhibit realistic agricultural behavior.

Overall, the benchmark contains very few missing values across all sensing modalities. Metadata, crop labels, yield statistics, geographic coordinates, and satellite observations are complete for essentially all samples. Small fractions of missing values remain only for MODIS land surface temperature observations, affecting less than 1\% of all timesteps, primarily due to unavailable observations within individual compositing windows.

Missing Sentinel-2 observations caused by cloud contamination were represented using a predefined placeholder value ($-9999$), preserving fixed-length temporal sequences. Approximately 12.2\% of Sentinel-2 observations correspond to these placeholders, enabling downstream models to distinguish unavailable observations without requiring variable-length sequences.


The physical ranges of all retrieved variables were further examined to verify consistency with their corresponding remote sensing products. Table~\ref{tab:value_ranges} summarizes the observed ranges for the principal variables included in the benchmark.



\begin{table}[t]
\centering
\scriptsize
\setlength{\tabcolsep}{4pt}
\caption{Observed value ranges for the principal variables included in AgroBench after preprocessing. Placeholder values ($-9999$) used for missing Sentinel-2 observations were excluded from the reported statistics.}
\label{tab:value_ranges}

\begin{tabular}{lcccc}
\toprule
Variable & Minimum & Median & Maximum & Missing \\
\midrule
Sentinel-2 Bands (10) & 0.00 & Band-dependent & 2.27 & 12.2\% (-9999)\\
Sentinel-1 VV (dB) & -52.6 & -11.49 & 16.60 & 0\% \\
Sentinel-1 VH (dB) & -75.4 & -19.11 & 8.65 & 0\% \\
LST Day ($^\circ$C) & -5.37 & 27.78 & 52.95 & 0.71\% NaN \\
LST Night ($^\circ$C) & -14.71 & 15.57 & 30.75 & 0.60\% NaN \\
Precipitation (mm) & 0.00 & 32.73 & 266.98 & 0\% \\
Elevation (m) & 5 & 286 & 2011 & 0\% \\
Yield & 6.1 & 64.3 & 999 & 0\% \\
\bottomrule
\end{tabular}
\end{table}

Overall, these analyses demonstrate that the proposed benchmark provides geographically diverse, physically consistent, and temporally aligned multimodal observations that are immediately suitable for large-scale spatiotemporal machine learning without requiring additional geospatial preprocessing.
\section{Benchmark Protocol and Baseline Results}
\label{sec:benchmark}

To facilitate standardized evaluation on AgroBench, we establish a reproducible benchmark for weakly supervised crop yield prediction. The benchmark evaluates representative baseline models using the same analysis-ready multimodal inputs, ensuring that performance differences primarily reflect modeling capability rather than data preparation.

\subsection{Benchmark Task Definition}
\label{subsec:task_definition}

The benchmark considers supervised crop yield prediction from multimodal spatiotemporal observations. Each sample comprises a fixed-length sequence of Sentinel-2, Sentinel-1, climatic, and terrain variables together with the corresponding county-level crop yield label.

For baseline evaluation, each temporal sequence was flattened into a single feature vector, enabling conventional machine learning models to operate directly on the benchmark.


\subsection{Experimental Protocol}
\label{subsec:protocol}

To evaluate temporal generalization, all baseline models were assessed using a Leave-One-Year-Out (LOYO) cross-validation protocol. In each evaluation fold, one growing season was held out exclusively for testing while the remaining years were used for training. This process was repeated for every year from 2017 through 2024, and the reported performance corresponds to the average across all eight folds.

LOYO prevents temporal information leakage and provides a realistic assessment of model generalization under inter-annual variability in environmental and climatic conditions. Performance was evaluated using the coefficient of determination ($R^2$), root mean squared error (RMSE), mean absolute error (MAE), and Pearson correlation coefficient ($r$).

Five representative regression baselines were evaluated: Linear Regression, Random Forest, XGBoost, LightGBM, and a Multilayer Perceptron (MLP). Default implementations with minimal hyperparameter tuning were used to emphasize the benchmark rather than model optimization.

\begin{table}[t]
\centering
\caption{Mean baseline performance under the Leave-One-Year-Out (LOYO) protocol across eight evaluation folds.}
\label{tab:baseline_results}
\begin{tabular}{lcccc}
\toprule
Model & $R^2$ & RMSE & MAE & Pearson $r$ \\
\midrule
Random Forest & 0.580 & 120.43 & 66.01 & 0.775 \\
LightGBM & 0.578 & 120.35 & 64.03 & 0.766 \\
XGBoost & 0.574 & 121.03 & 65.94 & 0.764 \\
Linear Regression & 0.075 & 176.61 & 125.29 & 0.343 \\
MLP & 0.009 & 182.47 & 100.92 & 0.476 \\
\bottomrule
\end{tabular}
\end{table}

\subsection{Benchmark Results}
\label{subsec:benchmark_results}

Table~\ref{tab:baseline_results} summarizes baseline performance under the LOYO protocol. Ensemble tree-based methods consistently outperformed Linear Regression and the MLP across all evaluation metrics.

Random Forest achieved the highest average $R^2$ (0.580) and Pearson correlation ($r=0.775$), while LightGBM produced the lowest MAE (64.03). XGBoost delivered comparable performance, indicating that ensemble tree methods effectively capture nonlinear relationships within AgroBench. In contrast, Linear Regression and the MLP performed substantially worse, suggesting that simple linear models and generic feed-forward networks are insufficient to exploit the benchmark's multimodal spatiotemporal structure. These baselines establish a reproducible reference point for future transformer-based, sequence, and foundation-model approaches that can explicitly exploit AgroBench's temporal and multimodal structure.

The comparable performance of the ensemble methods suggests that AgroBench contains informative predictive signals while leaving substantial headroom for models capable of learning richer temporal dependencies and multimodal interactions. Consequently, the benchmark provides a challenging and reproducible evaluation platform for future sequence models, transformers, and geospatial foundation models.

\section{Comparison with Existing Datasets}
\label{sec:comparison}

Table~\ref{tab:dataset_comparison} compares AgroBench with representative public datasets for agricultural and Earth observation research. Existing benchmarks primarily support crop classification, semantic segmentation, self-supervised representation learning, or subnational crop yield prediction. None combine multimodal Earth observation data with weakly supervised pixel-level crop yield labels in a reproducible benchmark.

\begin{table*}[t]
\centering
\small
\setlength{\tabcolsep}{4pt}
\renewcommand{\arraystretch}{1.15}
\caption{Comparison of AgroBench with representative public agricultural Earth observation datasets.}
\label{tab:dataset_comparison}

\begin{tabular*}{\textwidth}{@{\extracolsep{\fill}}l l l c c c c c c c}
\toprule
\textbf{Dataset} &
\textbf{Task} &
\textbf{Spatial} &
\textbf{S1} &
\textbf{S2} &
\textbf{Climate} &
\textbf{Yield} &
\textbf{Pixel} &
\textbf{Temporal} &
\textbf{Benchmark} \\
\midrule
CropHarvest & Classification & Field  & \checkmark & \checkmark & \checkmark & -- & \checkmark & \checkmark & \checkmark \\
SSL4EO-S12  & SSL            & Patch  & \checkmark & \checkmark & --          & -- & \checkmark & \checkmark & -- \\
PASTIS      & Segmentation   & Parcel & -- & \checkmark & -- & -- & \checkmark & \checkmark & \checkmark \\
Sen4AgriNet & Classification & Parcel & -- & \checkmark & -- & -- & \checkmark & \checkmark & \checkmark \\
SICKLE      & Multi-task     & Field  & \checkmark & \checkmark & -- & \checkmark & \checkmark & \checkmark & \checkmark \\
CY-Bench    & Yield          & Subnational & -- & \checkmark & \checkmark & \checkmark & -- & \checkmark & \checkmark \\
CropNet     & Yield          & County & -- & \checkmark & \checkmark & \checkmark & -- & \checkmark & \checkmark \\
\midrule
\textbf{AgroBench (Ours)}
& \textbf{Yield}
& \textbf{Pixel}
& \checkmark
& \checkmark
& \checkmark
& \checkmark
& \checkmark
& \checkmark
& \checkmark \\
\bottomrule
\end{tabular*}
\end{table*}

CropHarvest~\cite{cropharvest}, PASTIS~\cite{pastis}, Sen4AgriNet~\cite{sen4agrinet}, and SICKLE~\cite{sickle} were developed primarily for crop classification, semantic segmentation, or multi-task agricultural analysis, whereas SSL4EO-S12~\cite{ssl4eo} focuses on self-supervised representation learning from unlabeled Sentinel imagery. More recent yield-oriented datasets such as CropNet~\cite{cropnet} and CY-Bench~\cite{cybench} support crop yield prediction using county- or subnational-level supervisory signals. However, these benchmarks do not provide weakly supervised pixel-level crop yield labels suitable for learning spatially explicit representations. 

AgroBench complements these resources by combining county-level USDA yield statistics with temporally aligned multimodal Sentinel-1, Sentinel-2, climate, and terrain observations through a fully reproducible weak supervision pipeline. The resulting benchmark enables standardized evaluation of pixel-level crop yield prediction while remaining directly extensible to additional crops, regions, and growing seasons.
\section{Limitations and Future Directions}
\label{sec:limitations}

AgroBench inherits county-level USDA NASS yield statistics as supervisory labels, assigning a common yield value to all sampled crop pixels within a county-year. Although county-level labels provide scalable supervision, they do not
capture within-county yield variability. Consequently, AgroBench should
be viewed as a weakly supervised benchmark rather than a source of pixel-level ground-truth yield labels.

The current benchmark focuses on five major U.S. crops between 2017 and 2024 and evaluates conventional machine learning models using flattened feature representations to establish reproducible reference performance. Their inability to explicitly model temporal dependencies and multimodal interactions highlights AgroBench's potential as a benchmark for future sequence models, transformers, and geospatial foundation models.

The reproducible pipeline enables straightforward extension to additional crops, geographic regions, Earth observation products, and auxiliary data sources such as soil properties, crop simulation outputs, and management information. Beyond crop yield prediction, AgroBench also provides a foundation for related tasks including crop condition assessment, biomass estimation, stress detection, phenology modeling, and multimodal geospatial foundation model pretraining.
\section{Conclusion}
\label{sec:conclusion}

We presented AgroBench, a reproducible benchmark for weakly supervised pixel-level crop yield learning that integrates publicly available county-level agricultural statistics with multimodal Earth observation data. The benchmark contains over 13 million observations from more than 788,000 crop pixels spanning eight growing seasons for five major U.S. crops.

Along with the dataset, AgroBench provides a reproducible data generation pipeline, comprehensive dataset characterization, and standardized benchmark protocols with representative baseline models. By releasing the complete pipeline and evaluation framework, AgroBench establishes a reproducible benchmark for multimodal agricultural remote sensing, weakly supervised learning, and geospatial foundation models, enabling rigorous and reproducible comparison of future crop yield prediction methods.

\section*{Acknowledgments}

Figure~1 was created with assistance from generative AI–based
design tools and subsequently reviewed, corrected, and
manually refined by the authors to ensure that it accurately
represents the methodology and benchmark described in this
work.
\clearpage

\bibliography{refs}
\clearpage

\appendix

\begin{center}
    {\LARGE\bfseries Appendix}
\end{center}

\vspace{1em}

\section{Complete Geographic Coverage}
\label{app:coverage}

This appendix provides detailed information regarding the geographic
coverage of AgroBench. The benchmark was constructed using publicly
available county-level crop yield statistics reported by the USDA
National Agricultural Statistics Service (NASS). Only counties with
consistently reported yield statistics during the study period were
included. Consequently, the benchmark focuses on the principal U.S.
production regions for each crop while maintaining broad geographic
diversity.

AgroBench spans five major U.S. crops across multiple states and growing seasons. Since each crop is cultivated in distinct agroecological regions, the benchmark intentionally reflects real-world production patterns rather than enforcing uniform geographic coverage. 

Table~\ref{tab:state_coverage} lists the states represented for each
crop together with the corresponding temporal coverage and benchmark
size.

\begin{table*}[t]
\centering
\small
\caption{Detailed geographic coverage of AgroBench.}
\label{tab:state_coverage}

\begin{tabular}{lllll}
\toprule
Crop &
Years &
States Covered &
County-years &
Unique Pixels \\
\midrule

Corn &
2017--2024 &
IA, IL, IN, MI, MN, MO, OH, WI &
1713 &
313,598 \\

Soybeans &
2017--2024 &
IA, IL, IN, KS, LA, MI, MN, MO, MS, NE, WI &
2092 &
344,812 \\

Wheat &
2017--2023 &
CO, IN, KS, MT, NE, OH, OK, OR, SD, TX, WA &
776 &
126,534 \\

Cotton &
2017--2024 &
AL, GA, MS, NC, TX &
431 &
67,830 \\

Sorghum &
2017--2024 &
OK, TX &
95 &
14,246 \\

\bottomrule
\end{tabular}

\end{table*}

Corn and soybean primarily cover the U.S. Midwest, whereas wheat spans the Great Plains and Pacific Northwest. Cotton is concentrated in the southeastern and southern United States, while sorghum is included for Oklahoma and Texas, reflecting its major production regions.

\subsection{State Abbreviations}

Table~\ref{tab:state_names} provides the full names corresponding
to the abbreviations used throughout the benchmark.

\begin{table}[h]
\centering
\small
\caption{State abbreviations used in AgroBench.}
\label{tab:state_names}

\begin{tabular}{ll}
\toprule
Abbreviation & State \\
\midrule
AL & Alabama \\
CO & Colorado \\
GA & Georgia \\
IA & Iowa \\
IL & Illinois \\
IN & Indiana \\
KS & Kansas \\
LA & Louisiana \\
MI & Michigan \\
MN & Minnesota \\
MO & Missouri \\
MS & Mississippi \\
MT & Montana \\
NC & North Carolina \\
NE & Nebraska \\
OH & Ohio \\
OK & Oklahoma \\
OR & Oregon \\
SD & South Dakota \\
TX & Texas \\
WA & Washington \\
WI & Wisconsin \\
\bottomrule
\end{tabular}
\end{table}

\section{Temporal Coverage}
\label{app:temporal}

\begin{figure*}[t]
    \centering
    \includegraphics[width=0.9\textwidth]{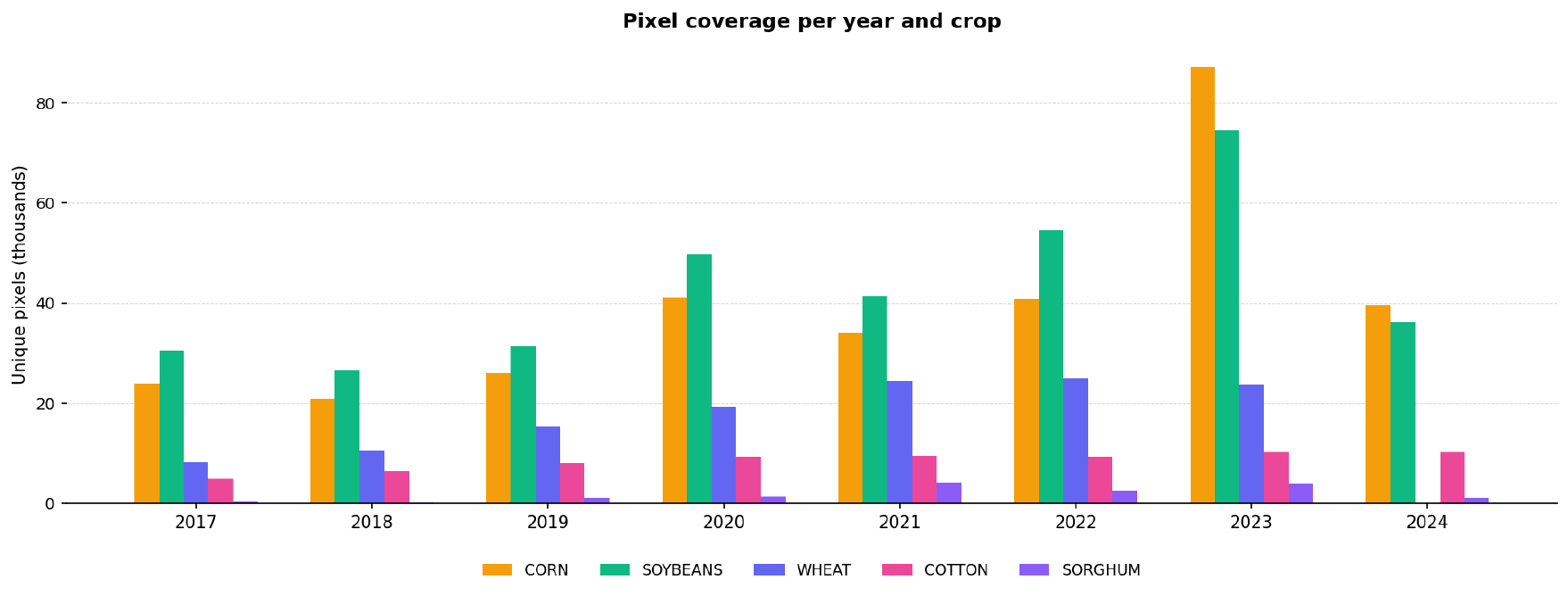}
    \caption{\textbf{Yearly distribution of unique crop pixels in AgroBench.}
    The figure shows the number of unique benchmark pixels retained for
    each crop from 2017--2024 after quality control and sampling. Corn
    and soybean constitute the largest portions of the benchmark across
    all years, while wheat, cotton, and sorghum exhibit crop-specific
    spatial coverage consistent with their major production regions.}
    \label{fig:pixel_coverage}
\end{figure*}

The benchmark spans eight growing seasons between 2017 and
2024. Figure~\ref{fig:pixel_coverage} illustrates the yearly
distribution of unique crop pixels retained for each crop after
quality control and sampling. The number of retained pixels varies
across years due to differences in crop extent, satellite data
availability, and quality filtering, while maintaining broad
coverage for all benchmark crops.

Overall, AgroBench maintains broad temporal coverage across
the study period while reflecting variations in satellite data
availability and crop-specific reporting. Sample counts
generally increase after 2017 as additional county-level yield
records become available.

Although observations remain available for 2024, several crops
contain fewer sampled pixels than previous years. This reduction
primarily reflects changes in the USDA Cropland Data Layer
(CDL) production workflow. Beginning in 2024, the CDL was
generated using Sentinel-2 imagery at 10\,m spatial resolution
rather than the previous Landsat-based workflow, producing
differences in crop delineation and consequently the number of
valid crop pixels identified during benchmark construction.

The broad temporal diversity captured by AgroBench enables
future models to evaluate robustness under varying climatic
conditions, management practices, and inter-annual agricultural
variability.

\section{Additional Data Quality Assessment}
\label{app:quality}

The analyses presented in the main paper summarize the overall
quality of AgroBench. This appendix provides additional
statistics describing missing values, physical consistency of
retrieved variables, and agronomic validation of the generated
time series.

\subsection{Missing Value Statistics}

Structural validation confirmed that metadata, crop identities,
county identifiers, geographic coordinates, and yield labels
are complete for essentially all benchmark samples.

Missing values are primarily associated with remote sensing
products affected by cloud contamination or sensor acquisition
frequency. Sentinel-2 observations unavailable after cloud
masking are represented using a placeholder value of
$-9999$ to preserve fixed-length temporal sequences.
Approximately 12.2\% of Sentinel-2 observations correspond
to these placeholders.

Missing values for MODIS daytime and nighttime land surface
temperature remain below 1\% of all observations, while
Sentinel-1 backscatter, precipitation, elevation, and yield
labels contain no missing values.

\subsection{Physical Consistency of Retrieved Variables}

All retrieved variables were examined to verify consistency
with the expected operating ranges of their respective Earth
observation products.

Sentinel-2 surface reflectance values remained within expected
reflectance ranges following preprocessing. Sentinel-1 VV and
VH backscatter exhibited distributions consistent with typical
agricultural scattering characteristics. MODIS land surface
temperatures and CHIRPS precipitation values agreed with
expected climatic conditions across the study region, while
SRTM elevation values closely matched the underlying digital
elevation model.

No systematic anomalies indicative of preprocessing or
retrieval failures were observed.

\subsection{Agronomic Validation Using NDVI}

To evaluate whether the generated temporal sequences preserve
meaningful vegetation dynamics, the Normalized Difference
Vegetation Index (NDVI) was computed from Sentinel-2
observations.

Crop-specific seasonal trajectories exhibited expected
phenological behavior. Corn and soybean generally reached the
highest peak-season NDVI values, whereas wheat exhibited an
earlier seasonal peak and cotton and sorghum displayed
distinct temporal growth patterns.

These observations provide additional evidence that the
temporal harmonization procedure preserves realistic crop
development throughout the growing season.

\section{Detailed Data Generation Pipeline}
\label{app:pipeline}

This appendix provides additional implementation details for each stage
of the AgroBench data generation pipeline. The benchmark is generated
entirely from publicly available geospatial datasets and agricultural
statistics using an automated and reproducible workflow. The modular
design of the pipeline enables straightforward extension to additional
crops, geographic regions, and growing seasons.

\subsection{County-Level Yield Statistics}

County-level crop yield statistics were obtained from the USDA National
Agricultural Statistics Service (NASS) Quick Stats database. Records
were collected for corn, soybeans, wheat, cotton, and sorghum between
2017 and 2024.

Several preprocessing steps were performed before benchmark generation.

\begin{itemize}

\item Aggregate records corresponding to ``OTHER COUNTIES'' were removed.

\item County identifiers were standardized using five-digit Federal
Information Processing Standards (FIPS) codes generated by combining
state and county ANSI identifiers.

\item Yield values were converted to numeric format while preserving
their original reporting units.

\item Counties lacking reliable yield observations were excluded.

\end{itemize}

The resulting dataset provides spatially consistent county-level weak
supervisory signals for benchmark construction.

\subsection{Crop Pixel Sampling}

Representative crop pixels were identified using the USDA Cropland Data
Layer (CDL). County boundaries were intersected with crop-specific CDL
masks to identify valid crop pixels corresponding to each county-year.

To maintain computational scalability while preserving broad spatial
coverage, at most 300 valid crop pixels were randomly sampled from each
county-year. Counties containing fewer than 300 crop pixels retained all
available observations.

Each sampled pixel inherited the county-level crop yield label while
retaining its own geographic location and multimodal temporal
observations, thereby forming the weakly supervised learning problem
introduced in the main paper.

\subsection{Earth Observation Retrieval}

Multimodal observations were collected from publicly available remote
sensing products.

The benchmark integrates

\begin{itemize}

\item Sentinel-2 Level-2A multispectral imagery (10 spectral bands),

\item Sentinel-1 Ground Range Detected (GRD) VV and VH backscatter,

\item MODIS daytime and nighttime land surface temperature,

\item CHIRPS precipitation,

\item SRTM digital elevation.

\end{itemize}

Cloud-contaminated Sentinel-2 observations were removed using the QA60
cloud mask. Images containing excessive cloud cover were discarded
before temporal compositing. Missing observations resulting from cloud
contamination were represented using the predefined placeholder value
$-9999$ in order to preserve fixed-length temporal sequences.

\subsection{Temporal Harmonization}

The incorporated Earth observation products differ substantially in
their revisit frequency and temporal availability.

To produce a unified representation, each growing season was divided
into consecutive 14-day intervals spanning April through October.
Within each interval,

\begin{itemize}

\item Sentinel-2 observations were median composited,

\item Sentinel-1 observations were aggregated,

\item MODIS land surface temperatures were averaged,

\item CHIRPS precipitation was accumulated,

\item Elevation remained constant.

\end{itemize}

This procedure produces a standardized sequence containing 15
temporally aligned observations for every sampled crop pixel.

\subsection{Final Benchmark Construction}

Following temporal harmonization, all sensing modalities were merged
into a unified benchmark representation.

Each record contains

\begin{itemize}

\item crop identifier,

\item county identifier,

\item geographic coordinates,

\item temporal metadata,

\item Sentinel-2 observations,

\item Sentinel-1 observations,

\item climatic variables,

\item terrain variables,

\item county-level crop yield label.

\end{itemize}

The benchmark is distributed in Apache Parquet format to enable
efficient storage, compression, and direct compatibility with modern
machine learning frameworks.

\section{Dataset Schema}
\label{app:schema}

Table~\ref{tab:features} summarizes the principal variables contained
within AgroBench together with their descriptions and units.

\begin{table*}[t]

\centering

\small

\caption{Feature schema of AgroBench.}

\label{tab:features}

\begin{tabular}{lll}

\toprule

Feature &
Description &
Units \\

\midrule

Latitude &
Pixel latitude &
Degrees \\

Longitude &
Pixel longitude &
Degrees \\

Year &
Growing season &
Integer \\

Crop &
Crop identity &
Categorical \\

County FIPS &
County identifier &
Integer \\

Yield &
County-level crop yield &
Crop-specific \\

B2--B8A, B11, B12 &
Sentinel-2 spectral bands &
Reflectance \\

VV &
Sentinel-1 VV polarization &
dB \\

VH &
Sentinel-1 VH polarization &
dB \\

LST Day &
MODIS daytime land surface temperature &
$^\circ$C \\

LST Night &
MODIS nighttime land surface temperature &
$^\circ$C \\

Precipitation &
CHIRPS accumulated precipitation &
mm \\

Elevation &
SRTM elevation &
m \\

Timestep &
14-day interval &
1--15 \\

\bottomrule

\end{tabular}

\end{table*}

\subsection{Sequence Representation}

Each crop pixel is represented as a fixed-length sequence containing 15
temporally aligned observations covering the primary growing season
between April and October.

Every timestep contains optical, radar, climatic, and terrain
variables describing the environmental conditions experienced by that
pixel during the corresponding 14-day interval. Together, these
observations form the multimodal input used throughout the benchmark.

\section{Dataset Organization}
\label{app:organization}

AgroBench is distributed using Apache Parquet files together with
metadata describing benchmark splits, crop identities, county
identifiers, and geographic locations.

The released repository is organized as follows.

\begin{verbatim}

AgroBench/

|-- data/

|   |-- corn/

|   |-- soybeans/

|   |-- wheat/

|   |-- cotton/

|   `-- sorghum/

|-- metadata/

|-- benchmark/

|-- scripts/

|-- documentation/

`-- README.md

\end{verbatim}

\subsection{Benchmark Files}

The benchmark includes

\begin{itemize}

\item multimodal crop-pixel observations,

\item county-level yield labels,

\item benchmark metadata,

\item evaluation splits,

\item preprocessing scripts,

\item data generation scripts,

\item baseline training scripts,

\item documentation describing benchmark usage.

\end{itemize}

\subsection{Reproducibility}

All preprocessing operations, benchmark generation procedures, and
baseline experiments are fully automated. Running the complete pipeline
using the released source code reproduces the benchmark from publicly
available datasets without requiring any manual intervention.

\section{Baseline Implementation Details}
\label{app:baseline}

All baseline models were implemented using widely used open-source
machine learning libraries.

Random Forest, Linear Regression, and MLP were implemented using
scikit-learn. XGBoost and LightGBM employed their official Python
implementations.

All experiments followed the Leave-One-Year-Out protocol described in
the main paper.

The benchmark includes five representative baseline models spanning
linear regression, ensemble learning, gradient boosting, and neural
network approaches.

\begin{itemize}

\item Linear Regression

\item Random Forest

\item XGBoost

\item LightGBM

\item Multi-Layer Perceptron (MLP)

\end{itemize}

These models were selected to provide diverse reference points while
maintaining modest computational requirements and straightforward
reproducibility.

All baseline models were evaluated using the Leave-One-Year-Out (LOYO)
protocol described in the main paper. During each evaluation fold, one
growing season was held out for testing while all remaining years were
used for training. Predictions were first generated at the pixel level
and subsequently aggregated to the county level by averaging all pixel
predictions belonging to the same county before computing evaluation
metrics.

Performance was evaluated using the coefficient of determination
($R^2$), root mean squared error (RMSE), mean absolute error (MAE), and
Pearson correlation coefficient. All reported results correspond to
county-level predictions following spatial aggregation of pixel-level
estimates.

\begin{table*}[t]
\centering
\small
\caption{Baseline model hyperparameters.}
\label{tab:baseline_hyperparameters}

\begin{tabular}{p{0.22\textwidth} p{0.70\textwidth}}
\toprule
\textbf{Model} & \textbf{Hyperparameters} \\
\midrule

Linear Regression &
StandardScaler + LinearRegression (default settings) \\

Random Forest &
300 trees; max\_features = sqrt; min\_samples\_leaf = 5; random\_state = 42 \\

XGBoost &
400 trees; max\_depth = 6; learning\_rate = 0.05; subsample = 0.8; colsample\_bytree = 0.8 \\

LightGBM &
400 trees; num\_leaves = 63; learning\_rate = 0.05; subsample = 0.8 \\

MLP &
Hidden layers = (256, 128, 64); learning rate = $10^{-3}$; maximum iterations = 500;
early stopping = True; validation fraction = 0.1 \\

\bottomrule
\end{tabular}
\end{table*}


\end{document}